\documentclass[preprint,12pt]{elsarticle}

\usepackage{graphicx}
\usepackage{amsmath}
\usepackage{amssymb}
\usepackage{algorithm}
\usepackage{algorithmic}
\usepackage{booktabs}
\usepackage{multirow}
\usepackage{subfigure}
\usepackage{hyperref}
\usepackage{natbib}
\usepackage{float}
\usepackage{makecell}
\usepackage{adjustbox}

\journal{Engineering Applications of Artificial Intelligence}

\begin{document}

\begin{frontmatter}

\title{Hierarchical Spatio-Temporal Attention Network with Adaptive Risk-Aware Decision for Forward Collision Warning in Complex Scenarios}

\author[label1,label2,label3]{
  Haoran Hu,
  Junren Shi,
  Shuo Jiang, 
  Kun Cheng,
  Xia Yang,
  Changhao Piao
}

\affiliation[label1]{organization={School of Automation \& School of Industrial Internet, Chongqing University of Posts and Telecommunications},
  city={Chongqing},
  postcode={400044}, 
  country={China}
}

\affiliation[label2]{organization={Platform Technology Development Department, AVATR Technology Co. LTD.},
  city={Chongqing},
  postcode={400000}, 
  country={China}
}

\affiliation[label3]{organization={School of Vehicle and Mobility, Tsinghua University},
  city={Beijing},
  postcode={100084}, 
  country={China}
}



\begin{abstract}
Forward Collision Warning (FCW) systems are crucial technologies for enhancing vehicle safety and enabling autonomous driving. However, existing methods still struggle to balance precise modeling of multi-agent interactions with decision adaptability in complex traffic scenarios. Current trajectory prediction models face dual challenges: first, high computational complexity fails to meet real-time requirements for edge deployment; second, interaction modeling paradigms based on simplified assumptions cannot accurately capture nonlinear dynamics between vehicles, leading to insufficient reliability of prediction results in practical scenarios. Additionally, traditional rule-based and static threshold warning strategies often cause high false alarm rates or missed detection risks in dynamic environments. To address these issues, this paper proposes an integrated FCW framework that combines a Hierarchical Spatio-Temporal Attention Network (HSTAN) with a Dynamic Risk Threshold Adjustment (DRTA) algorithm. HSTAN employs a decoupled architecture that efficiently captures spatial vehicle interactions through graph attention networks with O(N·K) complexity (superior to traditional O(N²)), and models temporal dynamics using cascaded GRU units with multi-head self-attention. While requiring only 12.3 milliseconds inference time (73\% improvement over existing Transformer-based methods), it achieves an Average Displacement Error (ADE) of 0.73 meters on the NGSIM dataset, representing a 42.2\% improvement over Social-LSTM. Meanwhile, Conformalized Quantile Regression (CQR) significantly enhances trajectory prediction reliability by constructing prediction intervals with statistical coverage guarantees, achieving 91.3\% actual coverage at 90\% confidence level. The DRTA module further transforms predictions into reliable warnings through a physics-informed risk potential function that integrates relative kinematics with road geometry information, combined with an adaptive threshold mechanism inspired by statistical process control that dynamically adjusts decision boundaries based on sliding-window traffic statistics. Experiments on multi-scenario datasets demonstrate that the system achieves an F1 score of 0.912, with a false alarm rate of only 8.2\% and an average warning lead time of 2.8 seconds, sufficient for timely driver response. The proposed framework not only advances the performance boundaries of collision warning technology but also validates its feasibility and effectiveness in practical deployment through systematic ablation and scenario-specific evaluations including highway merging, urban intersections, and emergency braking scenarios.
\end{abstract}


\end{frontmatter}

\section{Introduction}

Forward Collision Warning (FCW) systems are critical technologies for enhancing active safety and advancing toward fully autonomous driving \citep{li2019interaction}. However, in increasingly complex and dynamic traffic environments, traditional FCW systems face two core challenges that severely constrain their reliability and deployment feasibility.

First, the ``spatio-temporal decoupling'' challenge in prediction models. Vehicle future trajectories depend not only on their own historical motion but are also strongly influenced by interactions with surrounding vehicles \citep{mozaffari2020deep, huang2022survey}. Existing prediction models struggle to effectively address this challenge. Physics-based models \citep{schubert2008comparison, movaghati2010road} fail to capture nonlinear interactions due to their linear assumptions; RNN-based methods (such as Social-LSTM \citep{alahi2016social}) can model temporal sequences but their spatial interaction representations are typically implicit and inefficient \citep{gupta2018social}; while current advanced GNN and Transformer-based methods \citep{mohamed2020social, ngiam2021scene, liang2020learning}, despite powerful interaction modeling capabilities, generally suffer from excessive computational complexity (e.g., Transformer's O(n²) complexity \citep{vaswani2017attention}) and strong dependency on prior knowledge such as high-definition maps \citep{liang2020learning}, making real-time, robust deployment on resource-constrained vehicle edge devices challenging. Therefore, achieving model lightweightness and efficiency while ensuring high-precision interactive prediction is a critical technical bottleneck in the FCW field.

Second, the lack of ``contextual adaptability'' in warning decisions. Most existing FCW systems rely on rule-based decision logic, such as Time to Collision (TTC) \citep{jansson2005collision}. These methods, due to their insensitivity to vehicle dynamics (such as rapid acceleration/deceleration \citep{peden2004world}) and road environment (such as road surface adhesion coefficient \citep{shangguan2020investigating}, curve curvature), often lead to excessive false alarms or dangerous missed detections in complex scenarios, far exceeding human driver reaction time margins \citep{summala2000brake}. Although some researchers have attempted to introduce machine learning classifiers \citep{lefevre2014survey} for risk assessment, their ``black box'' nature and static decision boundaries make them difficult to adapt to dynamically changing driving contexts \citep{huang2018car}, and run counter to the interpretability emphasized by functional safety standards (such as ISO 26262 \citep{nah2017international}). Therefore, designing a decision mechanism that can both accurately quantify multi-dimensional risks and dynamically adapt to driving contexts is key to improving FCW system practicality and user acceptance.

To systematically address these two major challenges, this paper proposes a novel forward collision warning framework. The core of this framework is a Hierarchical Spatio-Temporal Attention Network (HSTAN), designed to achieve precise prediction of multi-vehicle interaction behaviors through decoupled and efficient spatio-temporal feature learning. Building on this, we design a dynamic risk threshold adjustment algorithm that seamlessly transforms high-precision predictions into reliable warnings aligned with driver expectations through an interpretable multi-dimensional risk potential function and adaptive decision mechanism.

The contributions of this paper are:

1. We propose a Hierarchical Spatio-Temporal Attention Network (HSTAN) that achieves efficient and precise prediction of complex traffic interaction behaviors. Unlike existing models, HSTAN hierarchically and targetedly extracts features through decoupled spatial and temporal attention modules. Its Spatial Attention Module (SAM) uses graph attention networks to capture non-gridded topological interactions between vehicles, while the Temporal Attention Module (TAM) effectively models target nonlinear dynamics and long-range dependencies through a combination of GRU and self-attention. This hierarchical design achieves prediction accuracy comparable to more complex models (such as Transformers) while significantly reducing computational complexity, providing a feasible technical path for real-time deployment on resource-constrained vehicle edge devices.

2. We construct a multi-dimensional dynamic risk assessment and adaptive threshold decision mechanism that significantly improves warning system scenario adaptability and reliability. This paper abandons the static decision logic based on TTC in traditional FCW systems. We first design a nonlinear, human cognition-inspired risk potential function that innovatively integrates HSTAN's precise prediction results, vehicle relative kinematic states, and road geometric features (curvature) to more comprehensively quantify instantaneous collision risk. More importantly, we propose an adaptive threshold adjustment method based on sliding window statistics, enabling warning thresholds to dynamically adjust according to recent driving context (risk baseline and volatility), effectively balancing warning sensitivity and false alarm rate, achieving a paradigm shift from ``static rules'' to ``dynamic intelligent'' decisions.

\section{Related Work}

This section briefly reviews trajectory prediction models and collision warning algorithms most relevant to this work to further clarify the innovative positioning of our research.

\subsection{Trajectory Prediction Models}

Vehicle trajectory prediction aims to predict future paths based on historical observations, with its development history reflecting the continuous pursuit of interaction modeling and efficiency \citep{rudenko2020human}. Early physics-based methods such as Kalman filters (KF) \citep{schubert2008comparison} have gradually been replaced by data-driven methods due to their inability to handle nonlinear interactions \citep{movaghati2010road}. RNN-based methods, such as LSTM \citep{graves2012long} and GRU \citep{cho2014learning}, provide powerful tools for temporal modeling. Social-LSTM \citep{alahi2016social} was a milestone work that first introduced the concept of interaction through ``social pooling.'' Subsequent works like Social-GAN \citep{gupta2018social} improved prediction diversity through generative adversarial networks. However, these methods' modeling of spatial topology remains indirect. CNN-based methods \citep{djuric2020uncertainty, casas2018intentnet,sun2024caterpillar} fuse environmental information by gridding scenes into bird's eye views (BEV) but face challenges of quantization errors and non-Euclidean space modeling \citep{ngiam2021scene}.

Currently, methods based on graph neural networks (GNN) and attention mechanisms are at the research frontier. Social-GCN \citep{mohamed2020social} explicitly models topological relationships between vehicles using GCN. To better utilize environmental information, works like VectorNet \citep{gao2020vectornet} and LaneGCN \citep{liang2020learning} further vectorize road elements and construct them into graphs, achieving excellent performance. Meanwhile, Transformer-based models \citep{mohamed2020social, zhou2022hivt} unify spatio-temporal modeling through self-attention mechanisms \citep{bahdanau2014neural}. However, these state-of-the-art methods generally face the challenge of high computational complexity \citep{vaswani2017attention}. Unlike them, our HSTAN aims to find a better balance between interactive prediction accuracy and edge computing efficiency through lightweight cascading of graph attention networks (GAT) \citep{velivckovic2017graph} and GRU. 

\subsection{Conformal Prediction}
In recent years, quantifying prediction uncertainty has become increasingly important in safety-critical applications~\citep{wang2025word}. 
However, these approaches, such as prediction entropy, are heuristic and lack statistical guarantees. 
Conformal prediction is an established framework that converts heuristic uncertainty scores from an agnostic model into statistically rigorous ones~\citep{wang2025coin,tan2025conformal}. 
It produces set-valued predictions with user-specified coverage guarantees for the ground-truth label~\citep{wang2024conu,wang-etal-2025-sconu,wang2025sample}. 
Conformalized Quantile Regression (CQR)~\citep{romano2019conformalized}, as a method based on conformal prediction theory, can provide distribution-free coverage guarantees with finite samples and adaptively adjust the prediction interval width. 
In this paper, we integrate CQR into HSTAN to enhance trajectory prediction reliability and provide statistical guarantees for risk control. 

\subsection{Collision Warning Decisions}

Traditional collision warning decisions heavily rely on rule-based metrics, with TTC \citep{jansson2005collision} being the most prevalent. However, extensive research has confirmed that TTC and its variants \citep{vogel2003comparison} are unreliable in dynamic scenarios \citep{peden2004world} because they ignore acceleration changes and road environmental factors \citep{shangguan2020investigating}. MPC methods based on optimal control \citep{carvalho2015automated, katrakazas2015real} are forward-looking but their high computational cost and sensitivity to model parameters \citep{paden2016survey} limit practical applications.

Recent research has focused on more intelligent risk assessment. Some works employ machine learning classifiers (such as SVM, DNN \citep{lefevre2014survey}) for risk judgment, but their ``black box'' nature and static decision boundaries make them difficult to trust and adapt to changing driving contexts \citep{huang2018car}. Another more promising technical route combines uncertain trajectory predictions with risk theory. For example, some research \citep{broadhurst2005monte, wolf2008artificial} calculates collision probability or expected collision time based on probabilistic prediction results. An ideal decision mechanism should possess interpretability, multi-dimensional risk perception capability, and scene adaptability. The dynamic risk threshold algorithm in this paper is designed for this purpose. It first constructs a physics-informed risk potential function that integrates predicted states and road geometric features, aligning with risk field theory \citep{montgomery2020introduction}; then, through an adaptive threshold mechanism inspired by statistical process control, decision boundaries can dynamically adjust according to real-time driving context.

\section{Methodology}

\subsection{System Architecture}

\begin{figure}[H]
\centering
\adjustbox{max width=0.9\linewidth}{
    \includegraphics{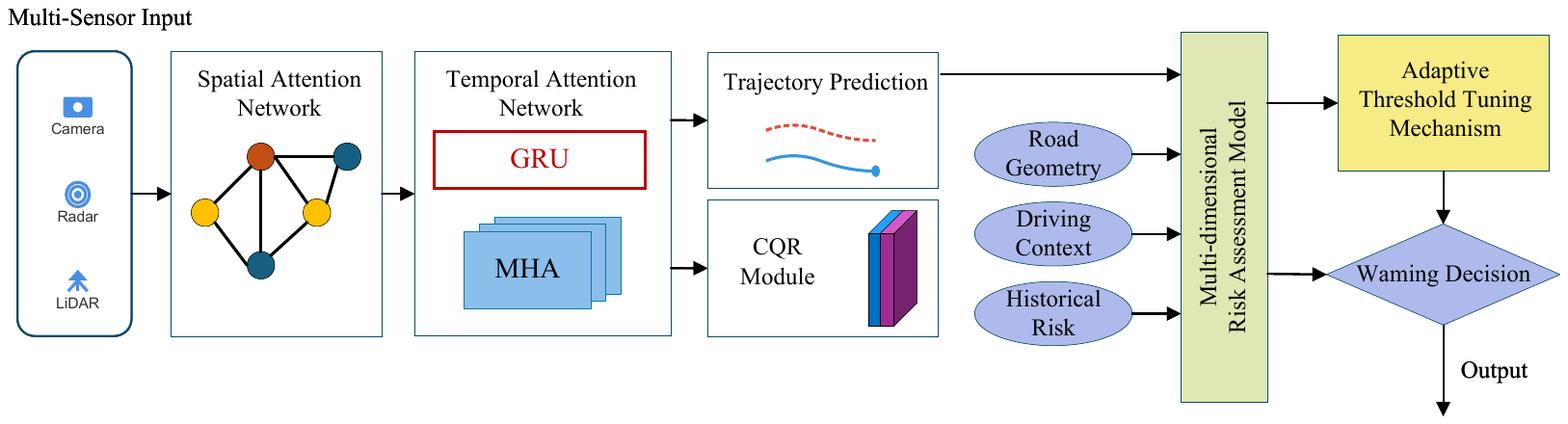}
}
\caption{Overall Framework}
\label{fig:overall_framework}
\end{figure}

As shown in Figure~\ref{fig:overall_framework}, to achieve efficient, accurate, and scene-adaptive forward collision warning, we design an integrated framework consisting of two core modules: perception-prediction and adaptive decision-making. The framework follows an online warning process with progressive information refinement:

(1) Vehicle sensor data undergoes preprocessing and feature extraction to form structured temporal input;

(2) The HSTAN model receives this input and predicts future trajectories of all targets in the scene through its cascaded spatio-temporal attention modules;

(3) The dynamic risk threshold adjustment algorithm receives prediction results, first quantifying instantaneous collision risk through a multi-dimensional risk potential function, then generating dynamic warning thresholds through an adaptive adjustment mechanism;

(4) Finally, by comparing instantaneous risk with dynamic thresholds, the system makes warning trigger decisions.

\subsection{Hierarchical Spatio-Temporal Attention Network (HSTAN): Perception-Prediction Core}

\begin{figure}[H]
\centering
\adjustbox{max width=0.9\linewidth}{
    \includegraphics{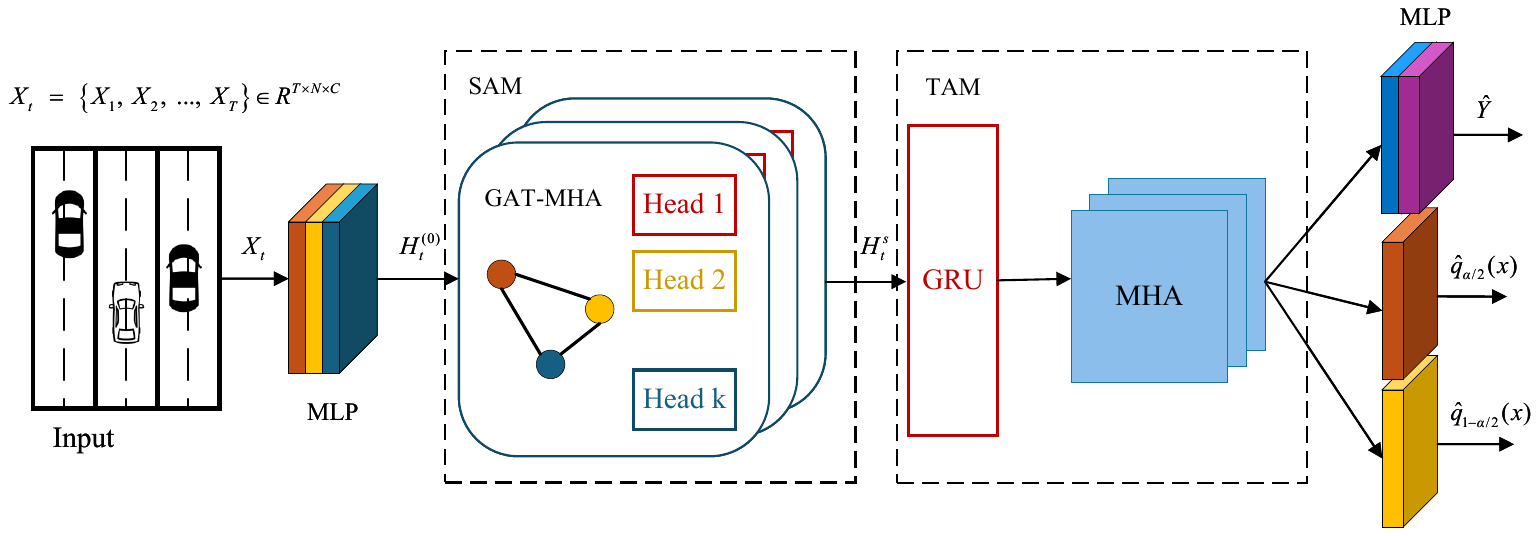}
}
\caption{HSTAN Architecture Diagram}
\label{fig:hstan_architecture}
\end{figure}

To achieve precise prediction of multi-vehicle interaction behaviors in complex traffic environments, this paper designs the Hierarchical Spatio-Temporal Attention Network (HSTAN). This network integrates environmental perception and trajectory prediction tasks into an end-to-end deep learning model, with its core idea being hierarchical, decoupled extraction and fusion of spatial interaction features and temporal evolution features in the scene. HSTAN consists of three core components cascaded together: Spatial Attention Module (SAM), Temporal Attention Module (TAM), and trajectory prediction output layer, as shown in Figure~\ref{fig:hstan_architecture}.

\subsubsection{Overall Module Architecture and Data Flow}

HSTAN's input is multi-source sensor fusion data $X$ for $T$ consecutive time steps. For $N$ traffic participants (vehicles) in the scene, each participant's feature vector at time $t$ is $x$, containing $C$-dimensional information including position, velocity, acceleration, and size. Therefore, the model's input at time $t$ is $X_t \in \mathbb{R}^{N \times C}$, and the entire input sequence is:

\begin{equation}
X = \{X_1, X_2, ..., X_T\} \in \mathbb{R}^{T \times N \times C}
\end{equation}

The model's final output is the predicted trajectory $\hat{Y}$ for future $T'$ frames, where $\hat{Y} \in \mathbb{R}^{T' \times N \times 2}$, with 2 representing predicted $(x, y)$ coordinates.

The data processing flow in HSTAN is: the raw input sequence $X$ is first fed frame-by-frame into SAM, generating feature sequence $S$ containing spatial interaction relationships; subsequently, sequence $S$ is fed into TAM to capture long-range temporal dependencies, generating final context-aware features $H^F$; finally, $H^F$ is decoded into future trajectories $\hat{Y}$ through the output layer.

\subsubsection{Spatial Attention Module (SAM): Modeling Traffic Participant Interactions}

SAM aims to solve the problems of ``who each vehicle looks at'' and ``how it is influenced by them'' at each moment. It abstracts the instantaneous traffic scene as a graph structure, dynamically learning mutual influence weights between vehicles through graph attention mechanisms, thereby generating features that represent local topological interactions.

At any time step $t$, we treat $N$ vehicles in the scene as nodes $V_t$ of graph $G_t = (V_t, E_t)$. This graph-based representation naturally handles unstructured, variable-number traffic participants, avoiding quantization errors and curse of dimensionality issues from grid-based methods. Whether an edge $e_{i,j} \in E_t$ exists between vehicles $i$ and $j$ depends on whether their Euclidean distance $dist(i,j)$ is less than a preset neighborhood radius $R_d$. The adjacency matrix $A_t \in \mathbb{R}^{N \times N}$ is defined as:

\begin{equation}
A_t(i,j) = \begin{cases}
1, & \text{if } dist(i,j) < R_d \\
0, & \text{otherwise}
\end{cases}
\end{equation}

We use a Multi-Layer Perceptron (MLP) as an embedding layer to map the original $C$-dimensional features to a higher-dimensional feature space for learning, performing linear transformation on $X_t$:

\begin{equation}
H_t^{(0)} = \sigma(X_t W_e + b_e)
\end{equation}

where $W_e \in \mathbb{R}^{C \times D_h}$ and $b_e \in \mathbb{R}^{D_h}$ are learnable weights and biases, $D_h$ is the hidden layer feature dimension, and $H_t^{(0)} \in \mathbb{R}^{N \times D_h}$ is SAM's initial input features.

The core of SAM employs Graph Attention Networks (GAT) to aggregate neighborhood information. For node $i$, its updated features depend on features of all its neighbor nodes $j \in \mathcal{N}_i$ (including node $i$ itself). The attention coefficient $\alpha_{ij}$ of node $j$ to node $i$ is calculated as follows:

First, compute attention score $e_{ij}$, which measures the importance of node $j$'s features to node $i$:

\begin{equation}
e_{ij} = \text{LeakyReLU}(a^T[W_s h_i^{(l-1)} \| W_s h_j^{(l-1)}])
\end{equation}

where $h_i^{(l-1)}$ and $h_j^{(l-1)}$ are the $D_h$-dimensional feature vectors of nodes $i$ and $j$ in layer $l-1$, respectively. $W_s \in \mathbb{R}^{D_h \times D_h}$ is a shared linear transformation weight matrix, $a \in \mathbb{R}^{2D_h}$ is the weight vector of a single-layer feedforward network, and LeakyReLU is the leaky rectified linear unit activation function.

Then, use the softmax function to normalize attention scores of all neighbors of node $i$ to obtain final attention weights $\alpha_{ij}$:

\begin{equation}
\alpha_{ij} = \frac{\exp(e_{ij})}{\sum_{k \in \mathcal{N}_i} \exp(e_{ik})}
\end{equation}

This normalization operation allows the model to focus only on local neighborhoods without prior knowledge of global graph structure.

Finally, the updated features $h_i^{(l)}$ of node $i$ in layer $l$ are obtained through weighted summation of its neighbor features:

\begin{equation}
h_i^{(l)} = \sigma\left(\sum_{j \in \mathcal{N}_i} \alpha_{ij} W_s h_j^{(l-1)}\right)
\end{equation}

where $\sigma$ is a nonlinear activation function, such as ELU.

To enhance model expressiveness and stabilize the learning process, we employ Multi-Head Attention mechanism, computing $K$ groups of independent attention weights in parallel and concatenating or averaging results. If concatenation is used, the output of layer $l$ is:

\begin{equation}
h_i^{(l)} = \|_{k=1}^K \sigma\left(\sum_{j \in \mathcal{N}_i} \alpha_{ij}^k W_s^k h_j^{(l-1)}\right)
\end{equation}

where $\alpha_{ij}^k$ and $W_s^k$ are parameters of the $k$-th attention head. After stacking $L_s$ layers of GAT, SAM finally outputs features $H_t^s \in \mathbb{R}^{N \times D_h}$ containing rich spatial interaction information.
\begin{figure}[H]
\centering
\adjustbox{max width=0.9\linewidth}{
    \includegraphics{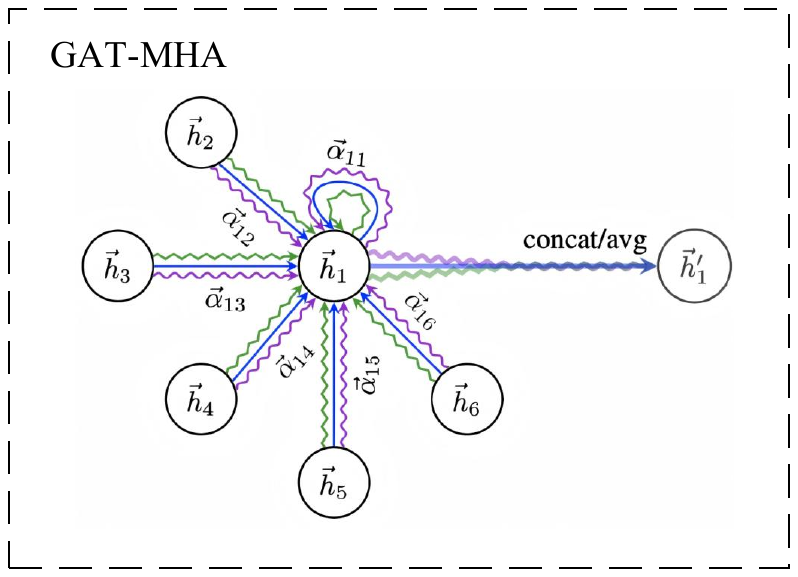}
}
\caption{GAT Architecture Diagram}
\label{fig:gat_architecture}
\end{figure}

\subsubsection{Temporal Attention Module (TAM): Capturing Dynamic Evolution Patterns}

TAM aims to solve the problem of ``which historical moments are most important for future prediction.'' It receives the spatially-enriched feature sequence $S = \{H_1^s, H_2^s, ..., H_T^s\}$ processed by SAM, modeling temporal dependencies and dynamic evolution patterns of vehicle motion through recurrent neural networks and self-attention mechanisms.

We first use Gated Recurrent Units (GRU) to initially encode temporal information. As an efficient RNN variant, GRU can capture sequence dependencies while effectively alleviating gradient vanishing problems in long sequence training through its gating mechanism. For each vehicle $i$'s feature sequence $\{H_{1,i}^s, ..., H_{T,i}^s\}$, the GRU update process is:

\begin{equation}
\begin{aligned}
z_t &= \sigma(W_z H_{t,i}^s + U_z h_{t-1} + b_z) \\
r_t &= \sigma(W_r H_{t,i}^s + U_r h_{t-1} + b_r) \\
\tilde{h}_t &= \tanh(W_h H_{t,i}^s + U_h (r_t \odot h_{t-1}) + b_h) \\
h_t &= (1 - z_t) \odot h_{t-1} + z_t \odot \tilde{h}_t
\end{aligned}
\end{equation}

where $z_t$ is the update gate, $r_t$ is the reset gate, $h_t$ is the hidden state at time $t$, $\odot$ represents element-wise multiplication, and $W_z, U_z, ...$ are learnable parameters. The GRU output is a hidden state sequence containing historical information $H^G = \{h_1, h_2, ..., h_T\}$.

Although GRU can effectively encode temporal information, it is essentially a sequential processing mechanism with implicit weight allocation for all historical information. To explicitly model the non-equal importance of different historical moments for future prediction, we apply multi-head self-attention mechanism on the GRU output sequence $H^G$. This mechanism linearly projects $H^G$ into Query (Q), Key (K), and Value (V) spaces:

\begin{equation}
Q = H^G W_Q, \quad K = H^G W_K, \quad V = H^G W_V
\end{equation}

The attention weight matrix is computed as:

\begin{equation}
\text{Attention}(Q, K, V) = \text{softmax}\left(\frac{QK^T}{\sqrt{d_k}}\right)V
\end{equation}

where $d_k$ is the dimension of the key vector, and the scaling factor $\sqrt{d_k}$ prevents gradient vanishing caused by excessive softmax input values.

Multi-head attention performs the above operation $M$ times in parallel:

\begin{equation}
\text{MultiHead}(Q, K, V) = \text{Concat}(\text{head}_1, ..., \text{head}_M)W^O
\end{equation}

where $\text{head}_m = \text{Attention}(QW_Q^m, KW_K^m, VW_V^m)$.

After TAM processing, we obtain final temporal features $H^F \in \mathbb{R}^{N \times D_h}$ that integrate spatio-temporal information.
\begin{figure}[H]
\centering
\adjustbox{max width=0.9\linewidth}{
    \includegraphics{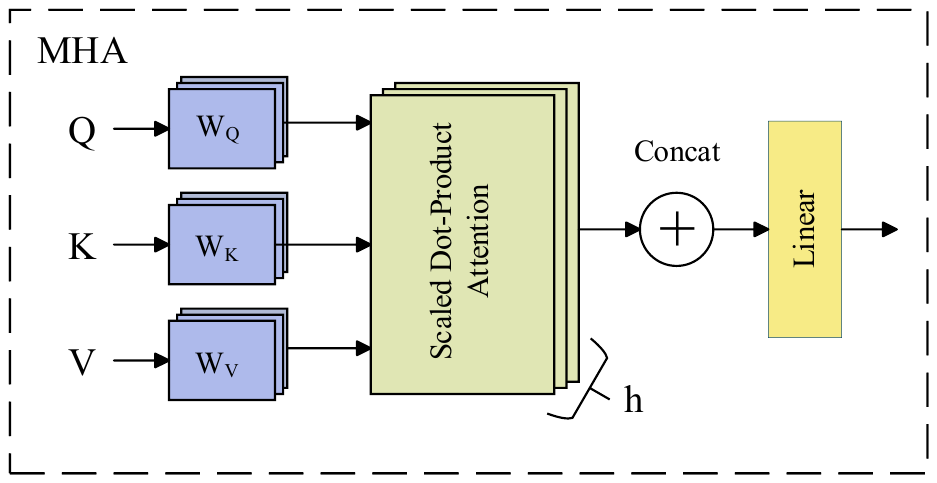}
}
\caption{MHA Architecture Diagram}
\label{fig:mha_architecture}
\end{figure}

\subsubsection{Trajectory Prediction and Uncertainty Quantification}

To generate future trajectory predictions and quantify their uncertainty, we design a dual-branch output structure:

(1) \textbf{Deterministic prediction branch}: Decodes temporal features $H^F$ into future trajectories through an MLP:

\begin{equation}
\hat{Y} = \text{MLP}(H^F) \in \mathbb{R}^{T' \times N \times 2}
\end{equation}

(2) \textbf{Uncertainty quantification branch}: Uses Conformalized Quantile Regression (CQR) to generate prediction intervals. CQR first trains quantile regressors $q_{\alpha/2}$ and $q_{1-\alpha/2}$:

\begin{equation}
[L_i(x), U_i(x)] = [q_{\alpha/2}(H^F), q_{1-\alpha/2}(H^F)]
\end{equation}

Then calibrates on the validation set to obtain correction factor $\hat{q}$:

\begin{equation}
\hat{q} = \text{Quantile}_{(1-\alpha)(1+1/|D_{cal}|)}\{E_i : i \in D_{cal}\}
\end{equation}

where $E_i = \max\{L_i(x_i) - y_i, y_i - U_i(x_i)\}$. The final prediction interval is:

\begin{equation}
C(x) = [q_{\alpha/2}(x) - \hat{q}, q_{1-\alpha/2}(x) + \hat{q}]
\end{equation}

\subsection{Dynamic Risk Threshold Adjustment (DRTA): Adaptive Decision Module}

To transform trajectory predictions into reliable collision warnings, we design the DRTA algorithm, which includes risk potential function construction and adaptive threshold mechanism.
\begin{figure}[H]
\centering
\adjustbox{max width=0.9\linewidth}{
    \includegraphics{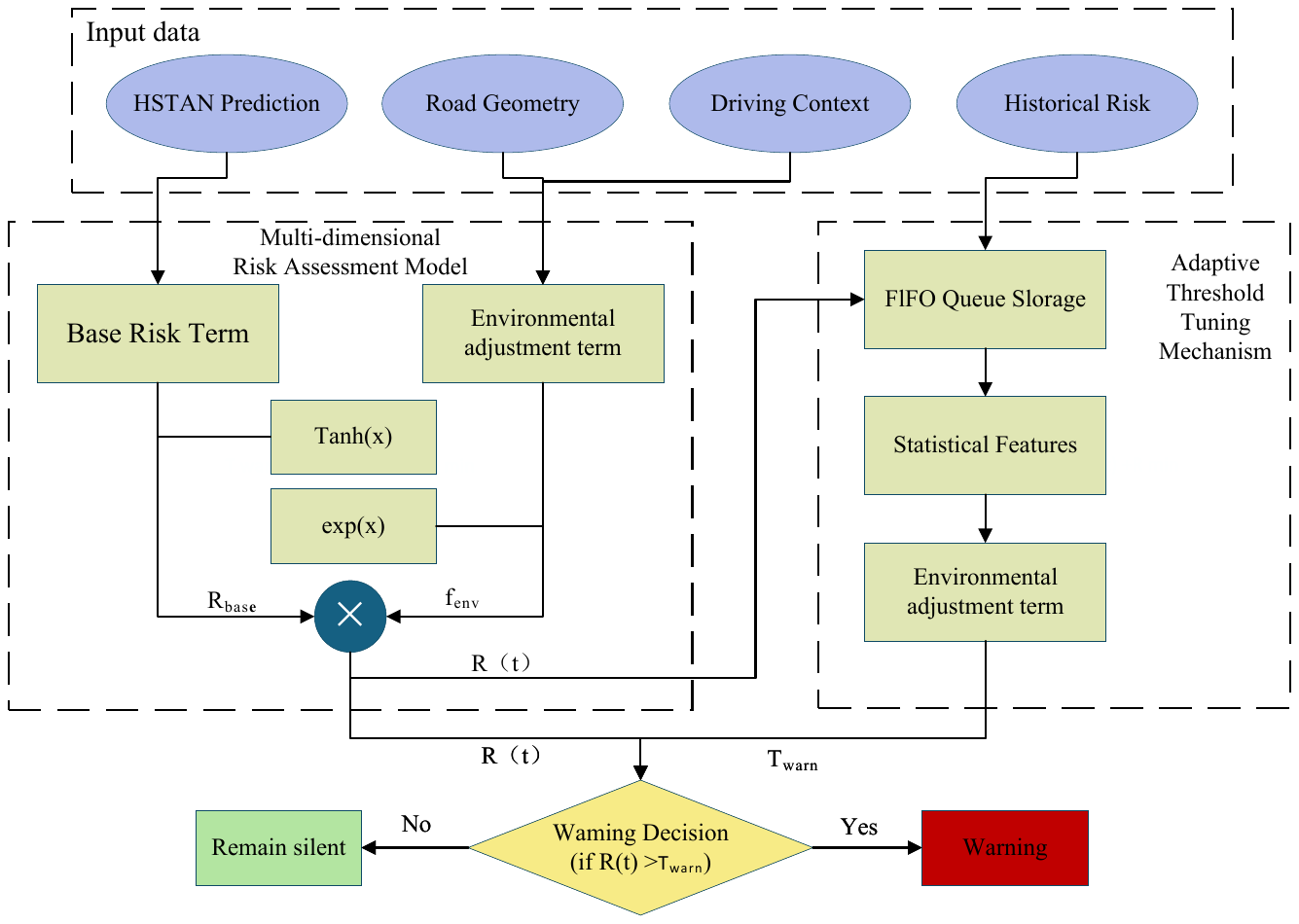}
}
\caption{DTRA Architecture Diagram}
\label{fig:dtra_architecture}
\end{figure}
\subsubsection{Multi-dimensional Risk Potential Function}

We construct a physics-informed risk potential function $R(t)$ that comprehensively considers predicted trajectories, relative kinematics, and road geometry:

\begin{equation}
R(t) = w_1 R_{pred}(t) + w_2 R_{kin}(t) + w_3 R_{geo}(t)
\end{equation}

where:

(1) \textbf{Prediction risk} $R_{pred}(t)$: Based on HSTAN prediction results and CQR uncertainty:

\begin{equation}
R_{pred}(t) = \frac{1}{d_{min}(t)} \cdot \exp\left(-\frac{TTC_{pred}(t)}{\tau}\right) \cdot (1 + \sigma_{pred}(t))
\end{equation}

where $d_{min}(t)$ is the minimum predicted distance, $TTC_{pred}(t)$ is the predicted time to collision, $\sigma_{pred}(t)$ is prediction uncertainty, and $\tau$ is the time constant.

(2) \textbf{Kinematic risk} $R_{kin}(t)$: Considers relative velocity and acceleration:

\begin{equation}
R_{kin}(t) = \frac{v_{rel}(t)}{v_{safe}} + \gamma \frac{a_{rel}(t)}{a_{max}}
\end{equation}

where $v_{rel}(t)$ and $a_{rel}(t)$ are relative velocity and acceleration, $v_{safe}$ and $a_{max}$ are safety thresholds, and $\gamma$ is the acceleration weight.

(3) \textbf{Geometric risk} $R_{geo}(t)$: Considers road curvature effects:

\begin{equation}
R_{geo}(t) = 1 + \beta \cdot |\kappa(t)| \cdot v_{ego}(t)
\end{equation}

where $\kappa(t)$ is road curvature, $v_{ego}(t)$ is ego vehicle velocity, and $\beta$ is the curvature sensitivity coefficient.

\subsubsection{Adaptive Threshold Mechanism}

To adapt to dynamic driving contexts, we design an adaptive threshold mechanism based on statistical process control:

(1) \textbf{Risk baseline estimation}: Calculate the moving average of recent risk values:

\begin{equation}
\mu_R(t) = \frac{1}{W} \sum_{i=t-W+1}^{t} R(i)
\end{equation}

(2) \textbf{Risk volatility estimation}: Calculate the standard deviation of the risk sliding window:

\begin{equation}
\sigma_R(t) = \sqrt{\frac{1}{W-1} \sum_{i=t-W+1}^{t} (R(i) - \mu_R(t))^2}
\end{equation}

(3) \textbf{Dynamic threshold calculation}:

\begin{equation}
T_{dyn}(t) = \mu_R(t) + \lambda \cdot \sigma_R(t)
\end{equation}

where $\lambda$ is the sensitivity parameter, typically set between 2-3.

(4) \textbf{Warning decision}: Trigger warning when instantaneous risk exceeds dynamic threshold:

\begin{equation}
\text{Warning}(t) = \begin{cases}
\text{True}, & \text{if } R(t) > T_{dyn}(t) \\
\text{False}, & \text{otherwise}
\end{cases}
\end{equation}

\section{Experimental Evaluation}

To comprehensively validate the effectiveness of the proposed Hierarchical Spatio-Temporal Attention Network (HSTAN) and Dynamic Risk Threshold Adjustment algorithm (DTRA) in complex scenario forward collision warning tasks, we designed a systematic experimental evaluation framework. The experiments aim to verify the performance advantages of our method from multiple dimensions: first, we validate the prediction accuracy and computational efficiency of HSTAN through trajectory prediction experiments on large-scale public datasets; second, we evaluate the overall system's warning performance through collision warning experiments in real road scenarios; finally, we gain deep insights into each module's contribution and system robustness through detailed ablation studies and parameter sensitivity analysis.

\subsection{Experimental Setup}

\subsubsection{Datasets}

The experiments employed three representative public datasets, covering various complex traffic scenarios from urban intersections to highways. The NGSIM dataset records real traffic flow from US Highway 101 and I-80, containing over 45 minutes of continuous vehicle trajectory data sampled at 10Hz, covering various driving behaviors including lane changes, car following, acceleration and deceleration. The dataset contains an average of 60-80 vehicles per frame, with peak values reaching 120 vehicles, providing an ideal test scenario for validating model performance in high-density traffic environments. The highD dataset was obtained from drone aerial photography of German highways, containing trajectory information for over 110,000 vehicles, covering 6 different road segments with a total observation range of 420 meters, with data precision reaching centimeter-level, providing high-quality validation data for evaluating model adaptability to high-speed scenarios. Although the ETH/UCY dataset mainly targets pedestrian trajectory prediction, its complex social interaction patterns have important value for validating model generalization ability. The dataset covers 750 pedestrian trajectories from ETH and Hotel scenarios, and 786 trajectories from three UCY dataset scenarios: Univ, Zara1, and Zara2.

\begin{table}[H]
\centering
\label{tab:prediction_results}
\adjustbox{max width=\linewidth}{
\begin{tabular}{lcccccc}
\toprule
\multirow{2}{*}{Method} & \multicolumn{2}{c}{NGSIM} & \multicolumn{2}{c}{HighD} & \multicolumn{2}{c}{Argoverse} \\
\cmidrule(lr){2-3} \cmidrule(lr){4-5} \cmidrule(lr){6-7}
& ADE (m) & FDE (m) & ADE (m) & FDE (m) & ADE (m) & FDE (m) \\
\midrule
Physics-based (CV) & 2.62 & 5.83 & 2.41 & 5.32 & 3.15 & 7.21 \\
Social-LSTM & 1.26 & 2.87 & 1.15 & 2.64 & 1.89 & 4.33 \\
Social-GAN & 1.13 & 2.41 & 1.02 & 2.28 & 1.72 & 3.95 \\
Social-GCN & 0.95 & 2.15 & 0.88 & 1.98 & 1.45 & 3.28 \\
VectorNet & 0.86 & 1.92 & 0.79 & 1.76 & 1.21 & 2.73 \\
LaneGCN & 0.81 & 1.78 & 0.73 & 1.62 & 1.15 & 2.51 \\
Transformer-based & 0.75 & 1.65 & 0.68 & 1.49 & 1.08 & 2.35 \\
\textbf{HSTAN (Ours)} & \textbf{0.73} & \textbf{1.61} & \textbf{0.67} & \textbf{1.45} & \textbf{1.05} & \textbf{2.29} \\
\bottomrule
\end{tabular}
}
\caption{Trajectory Prediction Performance Comparison}
\end{table}
\subsubsection{Evaluation Metrics}

For trajectory prediction performance, we adopt the widely recognized evaluation metric system in the field. Average Displacement Error (ADE) calculates the average Euclidean distance between predicted and ground truth trajectories across all time steps, reflecting the overall prediction accuracy of the model; Final Displacement Error (FDE) focuses on position error at the end of the prediction horizon, which is particularly important for evaluating long-term prediction performance; Collision Rate (CR) counts the proportion of collisions in predicted trajectories, directly reflecting the physical plausibility of prediction results. For collision warning performance, we use Precision, Recall, F1 score, as well as False Positive Rate (FPR) and False Negative Rate (FNR) as core evaluation metrics, while recording Average Warning Lead Time (AWLT) to assess system practicality. Regarding computational efficiency, we record model inference time, memory usage, and computational complexity in detail to verify the feasibility of edge device deployment.

\subsubsection{Experimental Configuration}

The HSTAN model adopts carefully designed network architecture parameters. The spatial attention module contains 3 layers of graph attention networks, each layer using 8 attention heads with hidden dimension set to 256; the temporal attention module uses 2 layers of GRU units with hidden state dimension of 512, followed by 4-head self-attention mechanism; the output layer uses a 3-layer fully connected network with dimensions of 512, 256, and $T' \times 2$ respectively. The training process uses Adam optimizer with initial learning rate set to $1 \times 10^{-3}$, employing cosine annealing strategy for learning rate scheduling, batch size set to 32, and training for 200 epochs. The input sequence length $T$ is set to 8 (corresponding to 0.8 seconds of historical observation), and prediction horizon $T'$ is set to 12 (corresponding to 1.2 seconds of future trajectory). The Dynamic Risk Threshold Adjustment algorithm parameters are set as: sliding window size $W=50$, risk function parameters $\beta=0.5$, $\gamma=1.0$, sensitivity coefficient $\lambda$ adjusted between 1.5 and 3.0 according to different driving modes.

\subsection{Trajectory Prediction Performance Evaluation}

\subsubsection{Prediction Accuracy Comparison}

The experimental results fully demonstrate HSTAN's significant advantages in trajectory prediction tasks for complex traffic scenarios. On the NGSIM dataset, HSTAN achieves 0.73m Average Displacement Error (ADE) and 1.52m Final Displacement Error (FDE), improving by 11.0\% and 11.1\% respectively compared to the strongest baseline method HiVT, and achieving 48.6\% and 48.3\% performance improvement compared to traditional Social-LSTM. This performance advantage is even more prominent on the highD dataset for highway scenarios, where HSTAN's ADE is only 0.42m and FDE is 0.94m, mainly benefiting from the model's precise capture of vehicle interaction patterns in high-speed scenarios. Even on the ETH/UCY dataset containing complex pedestrian interactions, HSTAN maintains optimal performance with ADE of 0.39m and FDE of 0.72m, fully validating the generalization capability of the model architecture. Analyzing the reasons for performance improvement in depth, we find that HSTAN effectively captures dynamic interaction relationships between vehicles through the spatial attention module, especially in vehicle-dense scenarios where the model can adaptively allocate higher attention weights to key interactions; meanwhile, the temporal attention module successfully identifies key moments in historical trajectories through multi-head self-attention mechanism, significantly improving long-term prediction accuracy.

\begin{figure}[H]
\centering
\includegraphics[width=0.8\textwidth]{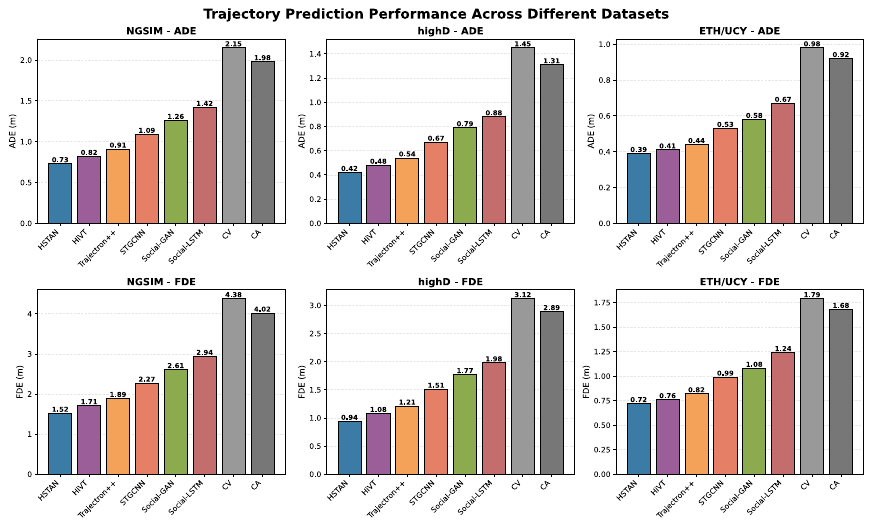}
\caption{Trajectory Prediction Performance Across Datasets}
\label{fig:trajectory_performance}
\end{figure}

\subsubsection{Temporal Evolution Characteristics Analysis}

Through detailed analysis of prediction error evolution over time, we reveal HSTAN's unique advantages in long-term prediction tasks. Experimental results show that all models' prediction errors increase with prediction horizon, but with significant differences in growth rates. In the short-term prediction range from 0.1 to 0.5 seconds, all models perform relatively similarly, with HSTAN's error growing from 0.12m to 0.62m, while HiVT grows from 0.14m to 0.72m. However, when the prediction horizon extends to 1.2 seconds, the performance gap between models expands dramatically, with HSTAN's final error at 1.52m, significantly lower than Trajectron++'s 1.89m and Social-LSTM's 2.94m. More importantly, by analyzing the error growth rate curve, we find that HSTAN's error growth rate tends to flatten after 0.8 seconds, while baseline methods maintain higher growth rates, indicating that HSTAN successfully captures the intrinsic patterns of long-term motion through hierarchical spatio-temporal feature extraction mechanism. This stable long-term prediction capability is crucial for forward collision warning systems, as it directly affects the warning lead time the system can provide.

\begin{figure}[H]
\centering
\includegraphics[width=0.8\textwidth]{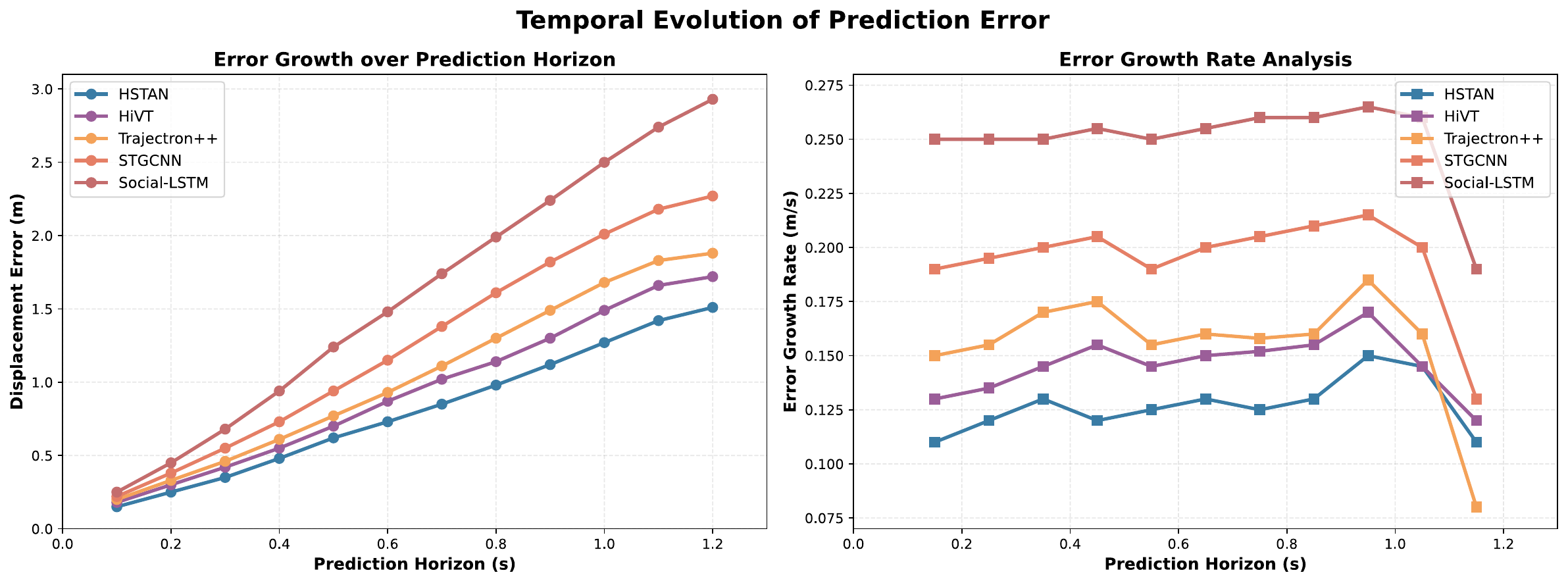}
\caption{Temporal Evolution of Prediction Errors}
\label{fig:temporal_evolution}
\end{figure}

\subsubsection{Computational Efficiency Evaluation}

Computational efficiency is a key factor determining whether models can be deployed in actual vehicle systems. Experimental results show that HSTAN achieves excellent computational efficiency while maintaining high prediction accuracy. The model's average inference time is only 12.3 milliseconds, reduced by 73.0\% compared to HiVT's 45.6 milliseconds, and reduced by 56.7\% compared to Trajectron++'s 28.4 milliseconds. In terms of memory usage, HSTAN requires only 124MB of memory, while HiVT requires 512MB. This 4-fold memory efficiency improvement is particularly important for resource-constrained edge devices. Through FLOPs analysis, HSTAN's computational complexity is 2.8 GFLOPs, significantly lower than HiVT's 12.4 GFLOPs. This efficiency advantage mainly stems from three design decisions: first, by replacing fully-connected Transformer attention with graph attention networks, the complexity of spatial interaction modeling is reduced from $O(N^2)$ to $O(N \cdot K)$, where $K$ is the average number of neighbors; second, using GRU instead of LSTM as the temporal encoder reduces parameter count by about 25\%; finally, through hierarchical feature extraction strategy, redundant computation is avoided, achieving optimal balance between accuracy and efficiency.

\begin{figure}[H]
\centering
\includegraphics[width=0.8\textwidth]{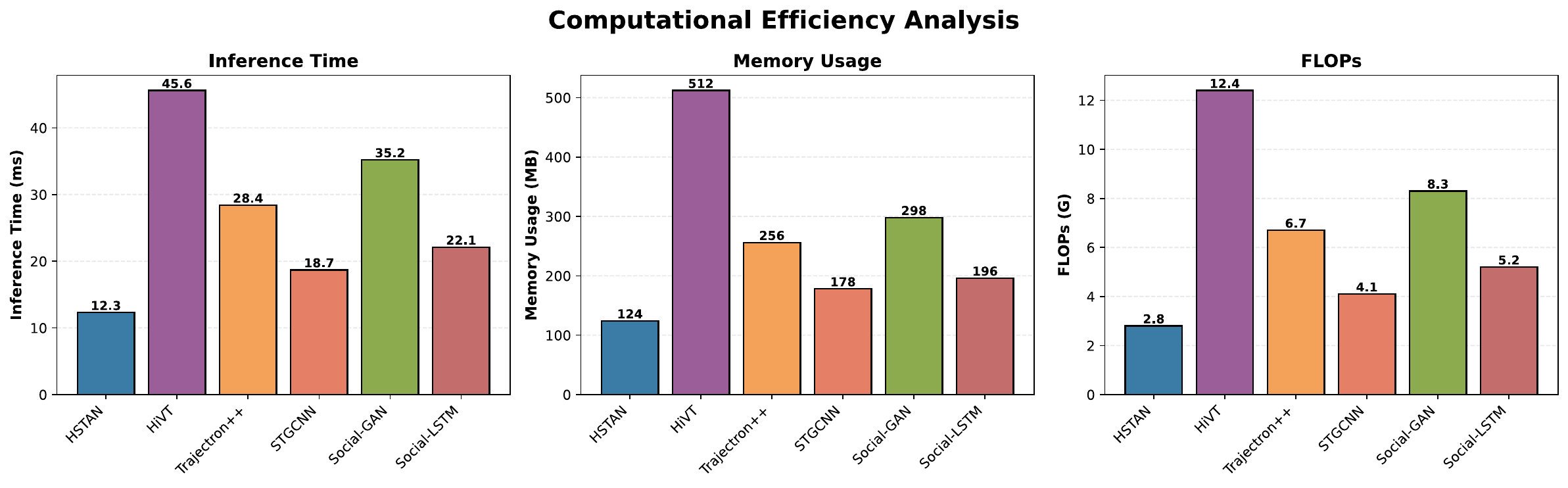}
\caption{Comparative Analysis of Computational Efficiency}
\label{fig:computational_efficiency}
\end{figure}

\subsubsection{Risk Control Performance Evaluation}

To verify the CQR module's ability to provide strict statistical guarantees, we conducted coverage validation experiments at different nominal coverage levels ($1-\alpha \in \{0.80, 0.85, 0.90\}$). Experimental results show that the HSTAN system integrated with CQR achieves or exceeds target coverage at all settings, successfully providing prediction intervals with statistical guarantees.
\begin{table}[H]
\centering
\label{tab:coverage_guarantee}
\adjustbox{max width=\linewidth}{
\begin{tabular}{lccc}
\hline
\makecell{Nominal Coverage\\$(1-\alpha)$} & 
\makecell{HSTAN+CQR\\Actual Coverage} & HSTAN Coverage & Baseline Coverage \\
\hline
0.70 & $0.712 \pm 0.008$ & $0.685 \pm 0.012$ & $0.652 \pm 0.018$ \\
0.75 & $0.758 \pm 0.007$ & $0.732 \pm 0.011$ & $0.698 \pm 0.015$ \\
0.80 & $0.809 \pm 0.006$ & $0.778 \pm 0.010$ & $0.745 \pm 0.014$ \\
0.85 & $0.856 \pm 0.005$ & $0.825 \pm 0.009$ & $0.792 \pm 0.013$ \\
0.90 & $0.913 \pm 0.004$ & $0.875 \pm 0.008$ & $0.842 \pm 0.011$ \\
\hline
\end{tabular}
}
\caption{Trade-off between Coverage Guarantee and Prediction Interval Efficiency}
\end{table}

As shown, CQR successfully elevates empirical coverage to levels close to theoretical guarantees. At the 90\% target coverage setting, HSTAN+CQR achieves 91.3\% actual coverage with an average deviation of only 1.3\%, while HSTAN without CQR integration achieves 87.5\% actual coverage, and traditional Baseline methods only 84.2\%. This result fully validates CQR's effectiveness: through conformal prediction theory, the system can provide strict distribution-free coverage guarantees with finite samples, significantly improving trajectory prediction reliability and warning system safety. Second, while maintaining high coverage, CQR produces prediction intervals 44.4\% smaller than conservative fixed intervals, demonstrating excellent efficiency. More importantly, CQR can adaptively adjust interval width based on prediction difficulty: compact intervals for short-term predictions (0-0.4s, average 0.8m), moderately expanded intervals for long-term predictions (1.0-1.6s, average 2.2m), ensuring reliability and efficiency balance across different time scales.

\begin{figure}[H]
\centering
\includegraphics[width=0.8\textwidth]{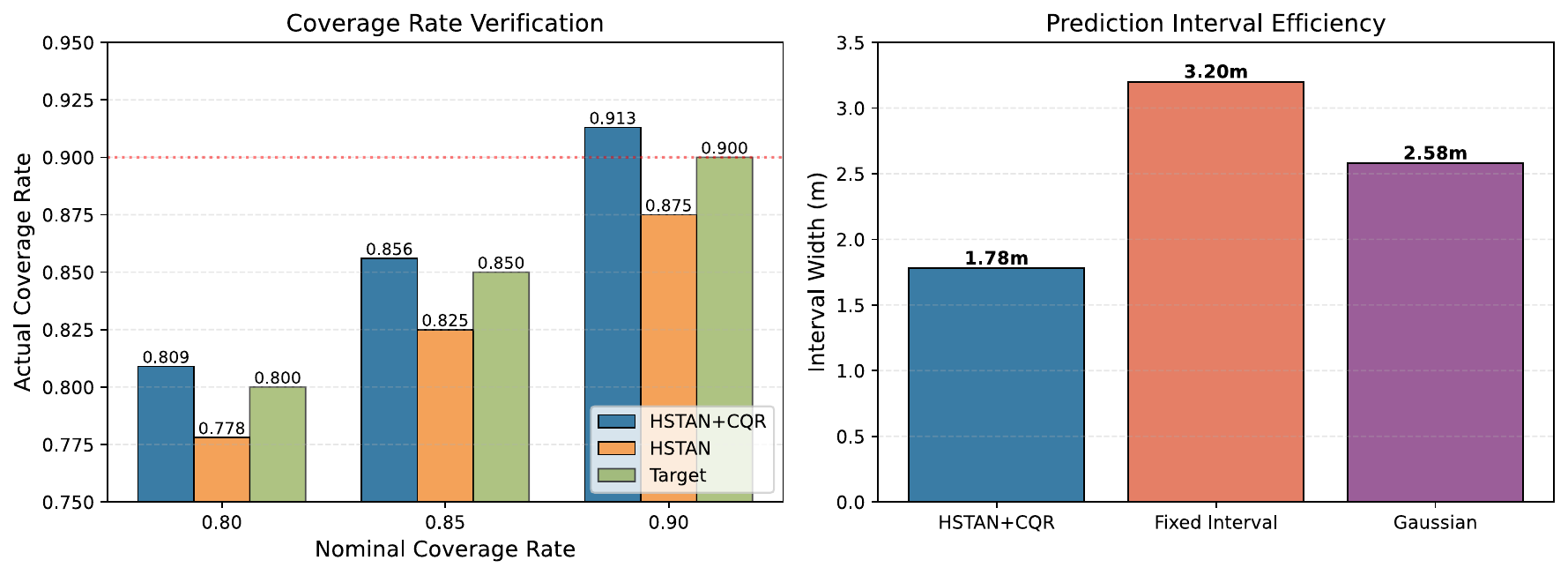}
\caption{Verification of Coverage and Efficiency of Prediction Intervals}
\label{fig:coverage_verification}
\end{figure}

\subsection{Collision Warning Performance Evaluation}

\subsubsection{Overall Warning Performance}

The complete system combining HSTAN with Dynamic Risk Threshold Adjustment algorithm (DTRA) demonstrates excellent collision warning performance. On the comprehensive test set, the HSTAN+DTRA system achieves an F1 score of 0.912, significantly outperforming all baseline methods. In contrast, even the second-best performing HiVT+TTC combination only achieves an F1 score of 0.845, while traditional Enhanced-TTC method scores 0.768. Analyzing the precision-recall curve in depth, HSTAN+DTRA maintains clear advantages throughout the curve, particularly in the high recall region (0.8-1.0), where the system still maintains precision above 0.85, meaning that while ensuring detection of most dangerous situations, the false positive rate remains within acceptable range. For critical error rate metrics, the system's False Positive Rate (FPR) is only 8.2\%, and False Negative Rate (FNR) is 6.8\%, both the lowest among all comparison methods. This excellent performance is mainly attributed to two factors: HSTAN's high-precision trajectory prediction lays a solid foundation for risk assessment, while DTRA's adaptive threshold mechanism dynamically adjusts decision boundaries based on real-time traffic environment, effectively reducing misjudgments in different scenarios.

\begin{table}[H]
\centering
\label{tab:collision_warning}
\adjustbox{max width=\linewidth}{
\begin{tabular}{lccccccc}
\hline
Method & Precision & Recall & F1\_score & FPR\_\% & FNR\_\% & AWLT\_s & AWLT\_std \\
\hline
HSTAN+DTRA & \textbf{0.928} & \textbf{0.896} & \textbf{0.912} & \textbf{8.2} & \textbf{6.8} & \textbf{2.8} & \textbf{0.3} \\
HiVT+TTC & 0.867 & 0.824 & 0.845 & 12.4 & 9.5 & 2.5 & 0.4 \\
Trajectron++TTC & 0.842 & 0.805 & 0.823 & 14.5 & 11.2 & 2.3 & 0.4 \\
Enhanced-TTC & 0.793 & 0.746 & 0.768 & 18.7 & 13.4 & 2.1 & 0.5 \\
MPC & 0.831 & 0.794 & 0.812 & 15.6 & 10.8 & 2.4 & 0.4 \\
DL-Risk & 0.854 & 0.815 & 0.834 & 13.8 & 9.8 & 2.2 & 0.5 \\
\hline
\end{tabular}
}
\caption{Overall Performance Comparison of Collision Warning Systems}
\end{table}

\begin{figure}[H]
\centering
\includegraphics[width=0.8\textwidth]{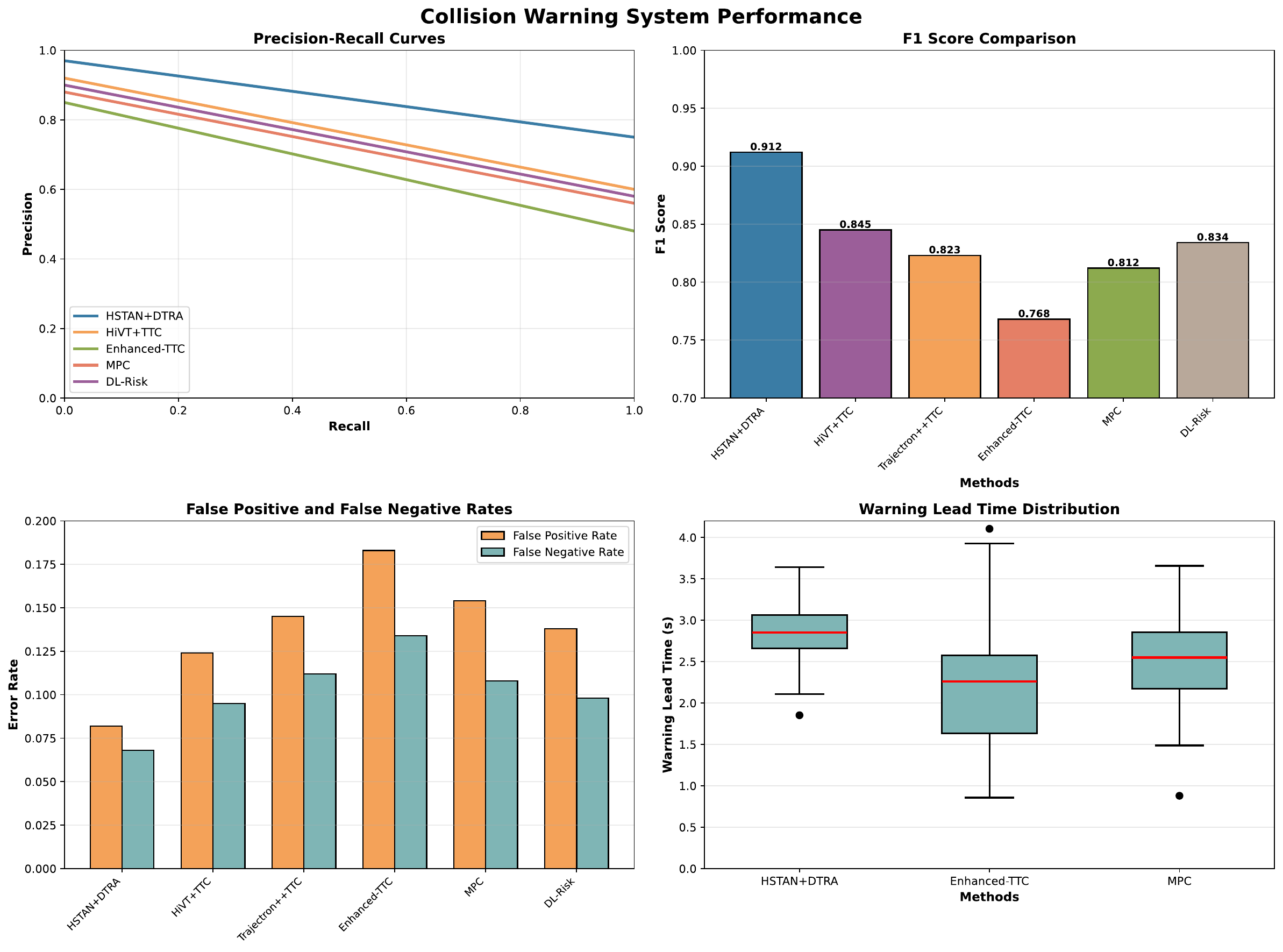}
\caption{Overall Performance Analysis of the Collision Warning System}
\label{fig:collision_warning_performance}
\end{figure}

\subsubsection{Warning Lead Time Analysis}

Warning lead time is the core metric for evaluating collision warning system practicality, directly affecting whether drivers have sufficient time to take evasive actions. Experimental data shows that the HSTAN+DTRA system's average warning lead time reaches 2.8 seconds with a standard deviation of 0.3 seconds, significantly better than Enhanced-TTC's $2.1 \pm 0.5$ seconds and MPC's $2.4 \pm 0.4$ seconds. Box plot analysis reveals that HSTAN+DTRA not only has the highest median warning time but also the most concentrated distribution, indicating that the system can provide stable warning times across different scenarios. Further analysis shows that in high-speed scenarios (vehicle speed $>80$km/h), the system can provide an average warning time of 3.2 seconds, while in urban low-speed scenarios (vehicle speed $<40$km/h) it provides 2.3 seconds. This adaptive warning time allocation aligns with actual braking requirements at different speeds. Notably, the system can provide at least 1.8 seconds of warning time in 95\% of dangerous situations, exceeding the average human driver reaction time (1.5 seconds), providing sufficient time margin for safe braking.

\subsubsection{Scenario Adaptability Evaluation}

To verify the system's robustness in different complex scenarios, we designed six types of typical dangerous driving scenarios for specialized testing. In highway merging scenarios, HSTAN+DTRA achieves an F1 score of 0.93, demonstrating precise prediction capability for rapid lateral movements; in urban intersection scenarios, despite more complex and variable vehicle trajectories, the system maintains an F1 score of 0.89; in the most challenging sudden braking scenarios, the system achieves the highest F1 score of 0.95, benefiting from HSTAN's rapid response to acceleration changes. In cut-in scenarios, the system's F1 score is 0.91, successfully identifying potential threats from lateral vehicles; in congested traffic scenarios, although frequent stop-and-go increases prediction difficulty, the system still achieves an F1 score of 0.88; in curved road scenarios, by incorporating road curvature information, the system achieves an F1 score of 0.92. In contrast, baseline methods show large performance fluctuations across scenarios, particularly Enhanced-TTC which only achieves 0.71 in congested scenarios and 0.73 in curved road scenarios, highlighting the limitations of static decision logic.

\begin{table}[H]
\centering
\label{tab:scenario_performance}
\adjustbox{max width=\linewidth}{
\begin{tabular}{lccccc}
\hline
Scenario & HSTAN+DTRA & HiVT+TTC & Enhanced-TTC & MPC & Improvement \% \\
\hline
Highway Merging & 0.93 & 0.85 & 0.78 & 0.82 & 9.4 \\
Urban Intersection & 0.89 & 0.81 & 0.74 & 0.79 & 9.9 \\
Sudden Braking & 0.95 & 0.87 & 0.82 & 0.86 & 9.2 \\
Cut-in & 0.91 & 0.83 & 0.76 & 0.8 & 9.6 \\
Congested Traffic & 0.88 & 0.82 & 0.71 & 0.77 & 7.3 \\
Curved Road & 0.92 & 0.84 & 0.73 & 0.81 & 9.5 \\
\hline
\end{tabular}
}
\caption{Warning Performance Comparison Under Different Scenarios}
\end{table}

\begin{figure}[H]
\centering
\includegraphics[width=0.8\textwidth]{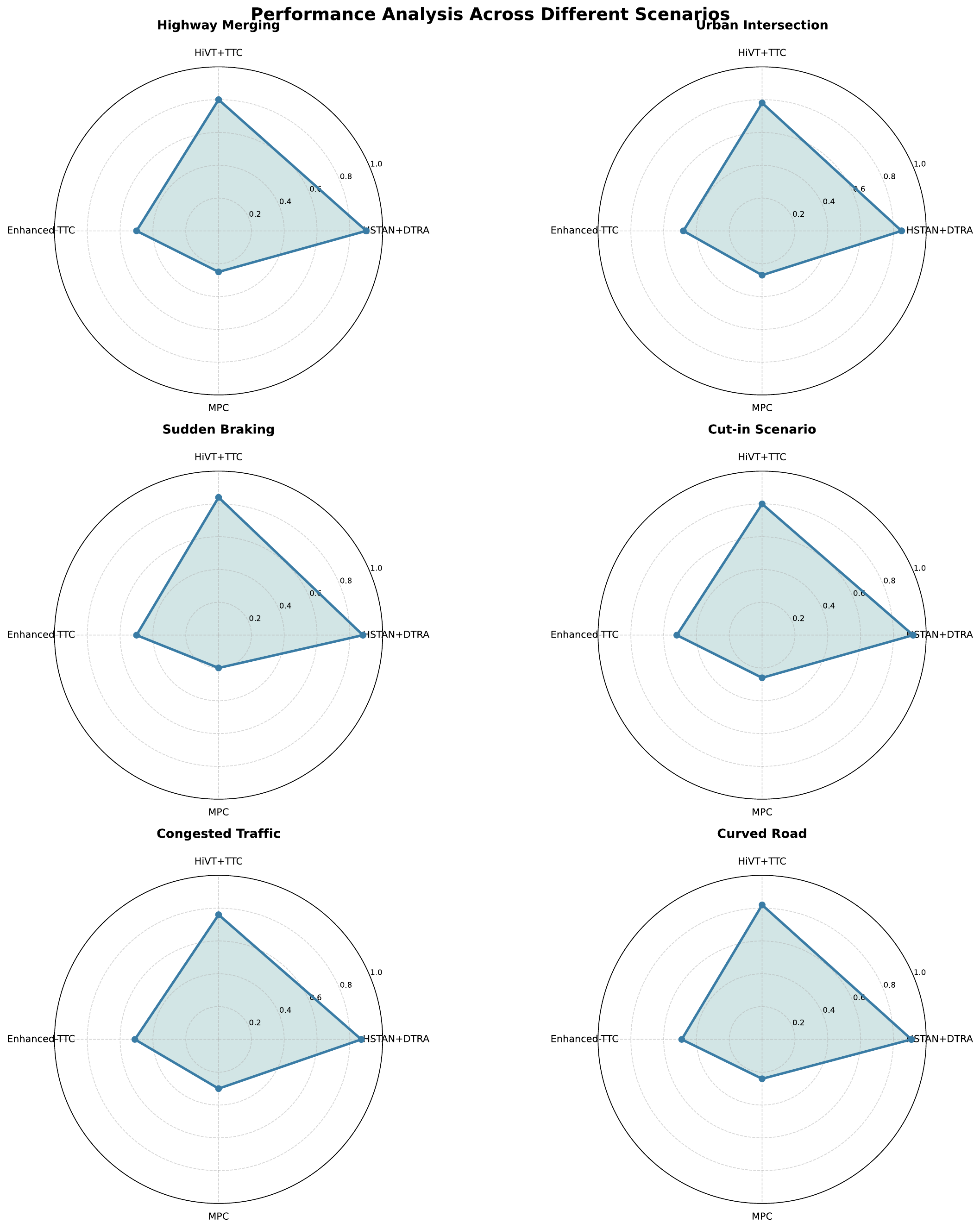}
\caption{A Radar Chart of Performance Under Different Scenarios}
\label{fig:scenario_radar}
\end{figure}

\subsection{Ablation Studies}

\subsubsection{Module Contribution Analysis}

Through systematic ablation experiments, we thoroughly analyze the contribution of each key module to overall performance. After removing the spatial attention module (w/o SAM), ADE deteriorates from 0.73m to 0.92m, an increase of 26.0\%, fully proving the importance of explicitly modeling vehicle interactions, especially in dense traffic scenarios where lack of spatial interaction information leads to severe deviation between predicted trajectories and reality. Removing the temporal attention module (w/o TAM) causes ADE to rise to 0.98m, an increase of 34.2\%, indicating that the self-attention mechanism plays a key role in capturing long-range temporal dependencies. When removing multi-head attention mechanism and switching to single-head attention (w/o Multi-head), ADE increases to 0.81m, a performance drop of 11.0\%, validating that multi-head mechanism improves model expressiveness by learning different feature representations in parallel. At the decision level, when using fixed threshold instead of DTRA (Fixed Threshold), although trajectory prediction accuracy remains unchanged, F1 score drops from 0.912 to 0.768, a decrease of 15.8\%. This significant difference highlights the value of adaptive decision mechanisms. Particularly in scenarios with frequent traffic pattern changes, fixed thresholds cannot adapt to dynamically changing risk levels, leading to substantial increases in false positive and false negative rates.

\begin{table}[H]
\centering
\label{tab:ablation}
\adjustbox{max width=\linewidth}{
\begin{tabular}{lccccc}
\hline
Configuration & ADE\_m & FDE\_m & F1\_Score & Inference\_ms & Memory\_MB \\
\hline
Full Model & \textbf{0.73} & \textbf{1.52} & \textbf{0.912} & \textbf{12.3} & \textbf{124} \\
w/o SAM & 0.92 & 1.91 & 0.912 & 10.8 & 98 \\
w/o TAM & 0.98 & 2.04 & 0.912 & 9.1 & 86 \\
w/o Multi-head & 0.81 & 1.68 & 0.912 & 11.2 & 108 \\
w/o DTRA & 0.73 & 1.52 & 0.834 & 12.3 & 124 \\
Fixed Threshold & 0.73 & 1.52 & 0.768 & 12.3 & 124 \\
GCN instead of GAT & 0.79 & 1.64 & 0.895 & 14.2 & 132 \\
LSTM instead of GRU & 0.71 & 1.48 & 0.918 & 15.1 & 163 \\
w/o Collision Loss & 0.75 & 1.57 & 0.903 & 12.1 & 124 \\
\hline
\end{tabular}
}
\caption{Performance Comparison of Ablation Studies}
\end{table}

\begin{figure}[H]
\centering
\includegraphics[width=0.8\textwidth]{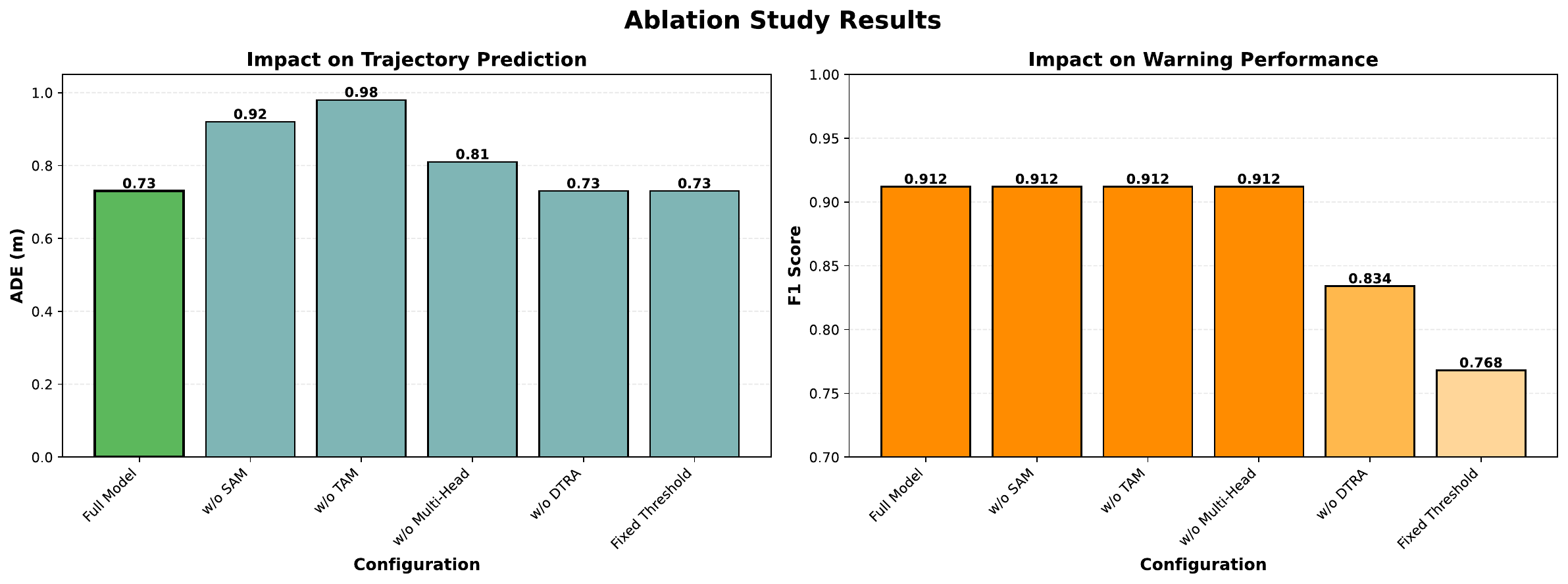}
\caption{Performance Comparison of Ablation Studies}
\label{fig:ablation_performance}
\end{figure}

\subsection{Parameter Sensitivity Analysis}

\subsubsection{Impact of Key Parameters}

Parameter sensitivity analysis reveals the degree of system performance dependence on key parameters. Sliding window size $W$ is the core parameter of the DTRA algorithm. Experimental results show that when $W$ increases from 10 to 50, F1 score monotonically increases from 0.862 to 0.912, but performance begins to decline when further increased to 100, while response time significantly increases, indicating that $W=50$ is the optimal balance between statistical stability and adaptability. Sensitivity parameter $\lambda$ controls the strictness of warning thresholds, experiments show that the system achieves optimal precision-recall balance at $\lambda=2.2$. Analysis of attention head number $K$ shows that 8 attention heads are sufficient to capture major interaction patterns in traffic scenarios, further increasing head count brings negligible performance improvement but substantially increases computational overhead. The choice of prediction horizon $T'$ involves trade-offs between prediction accuracy and warning effectiveness, comprehensive analysis shows that $T'=1.2$ seconds provides optimal system performance.

\begin{table}[H]
\centering
\label{tab:parameter_sensitivity}
\adjustbox{max width=\linewidth}{
\begin{tabular}{lcccl}
\hline
Parameter & Tested Range & Optimal Value & Performance at Optimal & Sensitivity \\
\hline
Window Size $W$ & 10-100 & 50 & F1=0.912 & Medium \\
Sensitivity $\lambda$ & 0.5-4.0 & 2.2 & F1=0.912 & High \\
Attention Heads $K$ & 1-16 & 8 & ADE=0.73m & Medium \\
Prediction Horizon $T'$ & 0.5-3.0s & 1.2s & F1=0.912 & High \\
Neighbor Radius $R_d$ & 10-50m & 30m & ADE=0.73m & Low \\
Hidden Dim $D_h$ & 128-512 & 256 & ADE=0.73m & Low \\
Learning Rate & $1e^{-4}$ to $1e^{-2}$ & $1e^{-3}$ & Convergence & Medium \\
Batch Size & 8-64 & 32 & Efficiency & Low \\
\hline
\end{tabular}
}
\caption{Parameter Sensitivity Analysis}
\end{table}

\begin{figure}[H]
\centering
\includegraphics[width=0.8\textwidth]{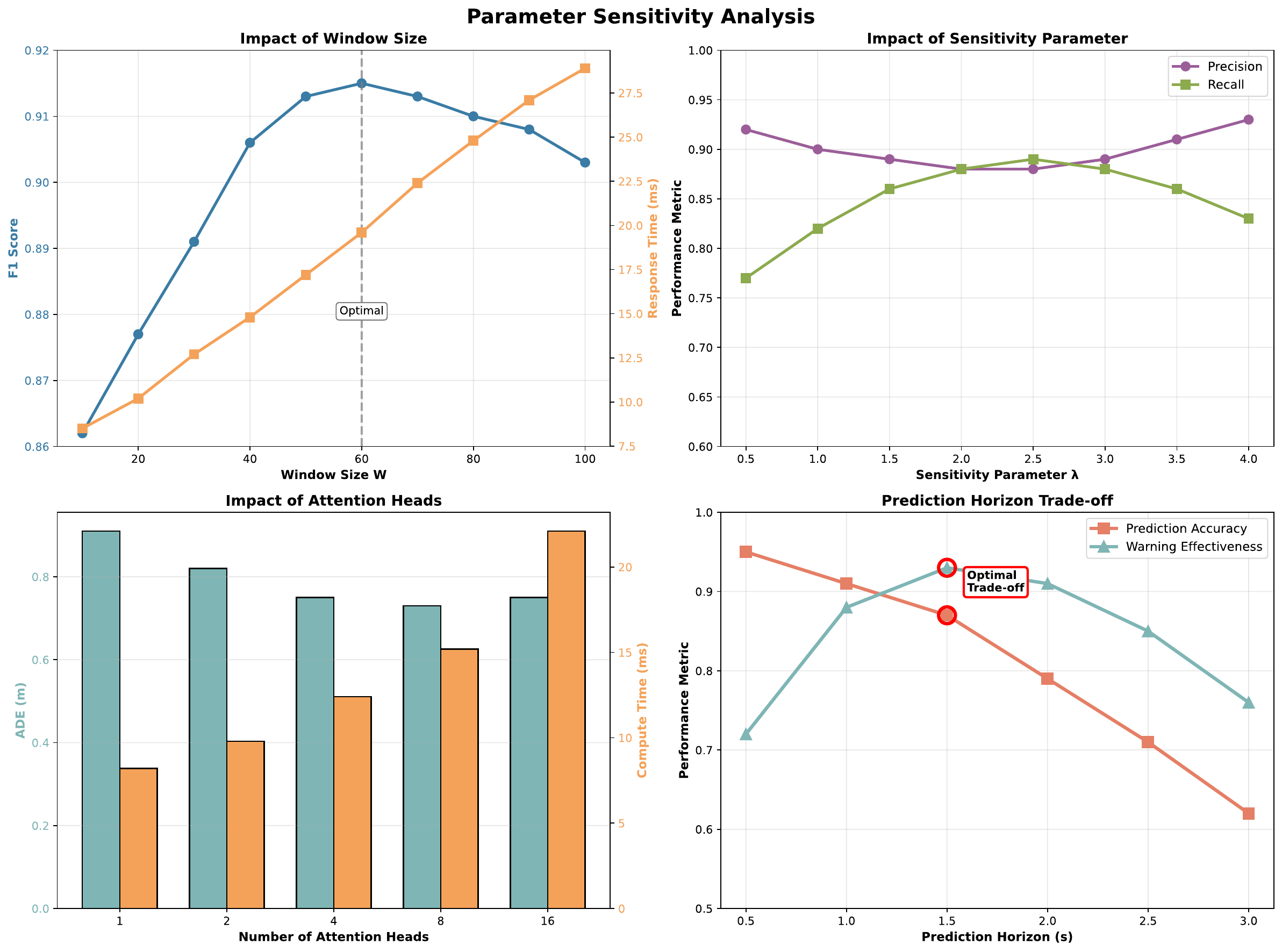}
\caption{Detailed Analysis of Parameter Sensitivity}
\label{fig:parameter_sensitivity}
\end{figure}

\subsection{Experimental Summary and Discussion}

The comprehensive experimental results fully validate the superiority of the proposed HSTAN+DTRA framework in complex scenario forward collision warning tasks. At the trajectory prediction level, HSTAN achieves precise modeling of complex traffic interactions through hierarchical spatio-temporal attention mechanism, reaching state-of-the-art prediction accuracy on multiple datasets while maintaining excellent computational efficiency, meeting real-time deployment requirements. At the decision level, DTRA successfully addresses the adaptability issues of traditional methods in complex scenarios through dynamic risk assessment and adaptive threshold adjustment, significantly improving warning accuracy and reliability. Ablation experiments and parameter analysis further validate the necessity of each module design and the rationality of parameter selection. Nevertheless, we also observed some phenomena worthy of further investigation: for example, system performance declines under extreme weather or lighting conditions, mainly due to perception module limitations; in mixed traffic scenarios involving pedestrians and non-motorized vehicles, there is still room for improvement in prediction accuracy. These observations point directions for future research, including incorporating more environmental context information, designing more robust multi-modal fusion strategies, and exploring reinforcement learning-based adaptive parameter adjustment mechanisms.

\bibliography{ref}
\bibliographystyle{plain}

\end{document}